\documentclass{article}
\usepackage{spconf,amsmath,graphicx,amsfonts, amssymb}
\usepackage{arydshln}
\usepackage{xcolor}
\usepackage{lipsum}
\usepackage{microtype}
\usepackage{graphicx}
\usepackage{subfigure}
\usepackage[hidelinks]{hyperref}       
\usepackage{booktabs} 
\usepackage{amsfonts}       
\usepackage{nicefrac}       
\usepackage{arydshln}

\usepackage{tikz}
\usepackage{pgfplots}

\usepackage[
backend=biber,
style=ieee,
citestyle=numeric-comp,
maxbibnames=3,
maxcitenames=3,
doi=false,isbn=false,url=false,eprint=false
]{biblatex}

\defbibheading{bibliography}[\refname]{}

\usepackage{cancel}

\makeatletter
\def\blfootnote{\xdef\@thefnmark{}\@footnotetext}
\makeatother

\title{Improving Massively Multilingual ASR With Auxiliary CTC Objectives}
\name{William Chen, Brian Yan, Jiatong Shi, Yifan Peng,  Soumi Maiti, Shinji Watanabe}
\address{Language Technologies Institute, Carnegie Mellon University, USA}
\begin{document}
\ninept
\maketitle
\begin{abstract}
Multilingual Automatic Speech Recognition (ASR) models have extended the usability of speech technologies to a wide variety of languages. With how many languages these models have to handle, however, a key to understanding their imbalanced performance across different languages is to examine if the model actually knows which language it should transcribe. In this paper, we introduce our work on improving performance on FLEURS, a 102-language open ASR benchmark, by conditioning the entire model on language identity (LID). We investigate techniques inspired from recent Connectionist Temporal Classification (CTC) studies to help the model handle the large number of languages, conditioning on the LID predictions of auxiliary tasks. Our experimental results demonstrate the effectiveness of our technique over standard CTC/Attention-based hybrid models. Furthermore, our state-of-the-art systems using 
self-supervised models with the Conformer architecture improve over the results of prior work on FLEURS by a relative 28.4\% CER. Trained models and reproducible recipes are available at \textcolor{blue}{\url{https://github.com/espnet/espnet/tree/master/egs2/fleurs/asr1}}.

\end{abstract}
\section{Introduction}
\label{sec:intro}
\vspace{-2mm}
Recent advancements in multilingual speech processing have shown great promise towards building speech systems for all, expanding language coverage beyond the high-resources \cite{watanabe2017language, adams2019massively, Pratap2020, Hou2020, li2020universal, conneau21_interspeech, liScaling, Yan2021DifferentiableAG, zhouConfigurable, bapna2022mslam, lu2022language, li2k_interspeech, ogayo22_interspeech, zhang2022streaming, bai2022joint, liLifeLong}.
In particular, practitioners have demonstrated that neural techniques are capable of large-scale cross-lingual representation learning by training multilingual automatic speech recognition (ASR) systems on massive private \cite{liLifeLong, liScaling, zhouConfigurable, Pratap2020} or public speech corpora \cite{gales14_babel, ardila-etal-2020-common, babu2021xls, chungW2v, bapna2022mslam, conneau2022fleurs}; however, these works demonstrate that performance still varies across languages. 

One of the inherent challenges in building a single system which can recognize many languages is the vast variability of phonology, grammar, and written scripts \cite{lewis2009ethnologue}.
Therefore, a key to understanding \textit{why} multilingual ASR systems exhibit certain errors is to examine whether the underlying model actually knows which language it should be transcribing -- in other words, if there was a correct language identification (LID).
To this end, systems that jointly model LID and ASR via multi-tasking \cite{watanabe2017Hybrid, zhouConfigurable, zhang2022streaming} offer one view to the inner-workings of the multilingual decision process.
However, we are interested in frameworks which more explicitly model LID as a dependency for ASR under the presumption that knowing the correct language of an utterance makes it easier to be transcribed. 

Therefore, in this work, we seek to build massively multilingual models which 1) condition transcription predictions on language identity likelihoods and 2) contribute our reproducible models and recipes which use publicly available data, with the broader objective of improving explain-ability and use-ability.

To achieve this, our work focuses on the FLEURS ASR dataset~\cite{conneau2022fleurs}. FLEURS contains 102 languages from across the globe, the typological diversity of which makes language identification a relevant component of transcription. Languages in FLEURS are also individually low-resourced: each language has only 7-10 hours of training data. This makes FLEURS a unique challenge that can help ASR progress to the long-tail of the world's $\sim$7000 languages.
We apply self-conditioned CTC \cite{interctc,,tjandra2020deja,sanabria2018hierarchical,nozaki2021relaxing, zhang2022intermediate, higuchihier, higuchiNar, yang2022improving, fujita2022multi}, which uses Connectionist Temporal Classification (CTC) models in intermediate encoder layers to condition subsequent layers on intermediate predictions, to Hybrid CTC/Attention architectures \cite{watanabe2017Hybrid} as a basis for our LID conditioning approach.
We then design intermediate LID targets of varying granularity and use these to examine the effect of conditioning our encoder-decoder models on LID predictions starting from early layers of the encoder.
Our proposed method, which allows early encoder layers to focus on LID while subsequent encoder and decoder layers focus on ASR, is beneficial compared to standard self-conditioning.
Together with self-supervised models and Conformer architectures, our state-of-the-art (SOTA) systems obtained 10.1 CER on FLEURS, a relative 28.4\% reduction over prior work.

\vspace{-1.5mm}
\section{Background}
\vspace{-1.5mm}
In this section, we discuss the CTC studies that we build upon to create our fully LID-conditioned model. These studies propose different  multi-task training methods \cite{watanabe2017Hybrid, interctc} and condition the encoder on intermediate predictions \cite{nozaki2021relaxing, higuchihier}.

\vspace{-2mm}
\subsection{Hybrid CTC/Attention}
\vspace{-1mm}
We use a Hybrid CTC/Attention architecture \cite{watanabe2017Hybrid} as our model foundation.
Let $X=(x_t \in \mathbb{R}^{D}|t=1,...,T)$ be a $T$-length input sequence based on $D$-dimensional acoustic feature $x_t$ and $Y=(y_s \in \mathcal{V}|s=1,...,S)$ be an $S$-length output sequence  with vocabulary $\mathcal{V}$.
CTC \cite{graves2006connectionist} optimizes a model to predict the monotonic alignment between $X$ and $Y$. It models the conditional probability of $P_{\mathsf{ctc}}(Y|X)$ as a series of latent sequences at each input frame. This latent sequence is obtained by introducing a blank token $\varnothing$ into $Y$, such that
$Z^{\mathsf{ctc}}=(z^{\mathsf{ctc}}_t \in \mathcal{V} \cup \{\varnothing\}|t=1,...,T)$.
\begin{equation}
    P_{\textsc{ctc}}(Y|X) \approx \sum_{Z^{\textsc{ctc}} \in F^{-1}(Y) } \prod_{t=1}^T P_{\textsc{ctc}}(z^{\textsc{ctc}}_t | X, \cancel{z_{1:t-1}})
    \label{eq:ctc_eq}
\end{equation}
Where $F^{-1}$ is the function of all latent sequences $Z^{\mathsf{ctc}}$ given $Y$. CTC operates with the assumption that only the observation $X$ is required to determine the latent state $z^{\mathsf{ctc}}_t$ at any given frame. The Hybrid CTC/Attention encoder first converts input $X$ into the hidden vector $\mathbf{h}$ in Equation (\ref{eq:enc_h}). The hidden vector is then used to obtain the frame-wise CTC posterior (Equation \ref{eq:P_ctc}) and token-wise attention posterior distributions of $X$ (Equation \ref{eq:P_att}).
\begin{align}
    \mathbf{h} &= \textsc{Enc}(X) \label{eq:enc_h}\\
    P_\textsc{CTC}(Z | X) &= \textsc{CTC}(\mathbf{h}) \label{eq:P_ctc}\\
    P_\text{Attn}(y_l | X, y_{1:l-1}) &= \textsc{Dec}(\mathbf{h}, y_{1:l-1}) \label{eq:P_att}
\end{align}
Combining Equations (\ref{eq:ctc_eq}, \ref{eq:P_ctc}, \ref{eq:P_att}) obtains the logarithmic linear combination of these posterior distributions over all frames and decoded tokens used to optimize the encoder-decoder network:
\begin{equation}
    \mathcal{L} = -(1-\lambda) \log P_\text{Attn}(Y|X) - \lambda \log P_\text{CTC}(Y|X)
    \label{loss_hybrid}
\end{equation}
\noindent Where $\lambda$ is the CTC weighting term. Hybrid CTC/Attention thus jointly optimizes a shared encoder with both CTC and attention losses, while the decoder is trained purely on the attention loss. 

\subsection{Intermediate CTC}
\label{sec:sc-ctc}
\noindent  InterCTC \cite{interctc,tjandra2020deja,sanabria2018hierarchical} was proposed to regularize the training of deep encoder networks by using the CTC loss of intermediate layers as part of a multi-task objective. The intermediate posterior distribution can be obtained in a manner similar to Equation (\ref{eq:P_ctc}), with the hidden vector $\mathbf{h}^\text{int}$ of an intermediate encoder layer.
\begin{align}
    P_\text{CTC}^\text{int}(Z^\text{int} | X) &= \textsc{CTC}^\text{int}(\mathbf{h}^\text{int}) \label{eq:P_inter}
\end{align}
The log-likelihood of Equation (\ref{eq:P_inter}) can then be used as the objective function of the intermediate layer. Self-conditioned CTC (SC-CTC) \cite{nozaki2021relaxing} also uses this intermediate output to condition the encoder by passing intermediate outputs to the next layer. 
The normalized hidden representation of the intermediate layer $\mathbf{h}^\text{int}$ is summed with a linear projection of the intermediate posterior distribution $Z^{\text{int}}$ to the hidden dimension, and input into the next encoder layer (Equation~\ref{eq:sc_add}). 
\begin{equation}
    \mathbf{h} = \textsc{Enc}^{\text{fin}}(\textsc{Nrm}(\mathbf{h^\text{int}}) + \textsc{Lin}(Z^\text{int}))) \label{eq:sc_add}
\end{equation}
\begin{equation}
    P_\text{CTC}(Z | X, Z^{\text{int}}) = \textsc{CTC}(\mathbf{h}) \label{eq:sc_ctc}
\end{equation}

Equation (\ref{eq:sc_add}) is recursively applied for each intermediate layer, until the output of the final layer is passed into Equation (\ref{eq:sc_ctc}). This allows the CTC posterior distribution of the entire encoder network to be conditioned on the intermediate predictions $Z^{\text{int}}$.

\vspace{-1.5mm}
\section{Proposed method}
\vspace{-1.5mm}
In this section, we propose an auxiliary CTC objective such that early encoder layers can focus on language identification, and the transcription predictions of later layers can be conditioned on the predicted language. We also define a hierarchy of CTC objectives to take advantage of both LID and self-conditioning, and frame it within a Hybrid CTC/Attention setup.

\subsection{Explicit multilingual conditioning}
\label{sec: lid-cond}

While SC-CTC's conditioning on early predictions benefits non-autoregressive models \cite{nozaki2021relaxing, higuchiNar}, it could be a drawback in the auto-regressive setting by conditioning later encoder layers on the noisy transcription predictions of earlier layers. We want to condition encoder layers on the LID without these noisy early predictions. To accomplish this, we propose the following method: train intermediate layers to only classify the spoken language and propagate their predictions to future layers via self-conditioning. We first introduce the latent variable $I$ to Equation (\ref{eq:ctc_eq}), where $I$ represents the intermediate LID predictions, such that Equation (\ref{eq:ctc_eq}) is modified as follows:
\begin{align}
     P_{\textsc{ctc}} (Y|X) &= \sum_{I \in \mathcal{I}} P_{\textsc{ctc}}(Y | I,X) P_{\textsc{ctc}} (I|X)
    \label{eq:lid_latent_2}
\end{align}
The formulation of Equation (\ref{eq:lid_latent_2}) can be realized by modifying the intermediate CTC network (Equations \ref{eq:sc_add} and \ref{eq:sc_ctc}) to predict the LID instead of transcriptions as follows:
\begin{equation}
    \mathbf{h} = \textsc{Enc}^{\text{fin}}(\textsc{Nrm}(\mathbf{h^\text{lid}}) + \textsc{Lin}(Z^\text{lid}))) \label{eq:lid_add}
\end{equation}
\begin{equation}
    P_\text{CTC}(Z | X, Z^{\text{lid}}) = \textsc{CTC}(\mathbf{h}) \label{eq:lid_ctc}
\end{equation}

\noindent This allows the encoder to condition its predictions on the LID. We then define an auxiliary CTC task with Equations (\ref{eq:ctc_eq}) and (\ref{eq:lid_latent_2}), where the model's intermediate layers attempt to predict the language,  $I$.

\begin{equation}
    \mathcal{L}^{\mathsf{lid}} = -\log P_{\textsc{ctc}}(I|X)
    \label{eq:lid_loss}
\end{equation}

\begin{table}[tb]
\caption{Comparison of different labels in the multi-task framework (Sec. \ref{sec: lid-cond}). In LID$_{\mathsf{tok}}$, all tokens are replaced with LIDs, while LID$_{\mathsf{utt}}$ only retains a single LID label.}
\label{tab:labels}
\centering
\begin{tabular}{l|l}
\hline
Task & Label\\
\hline
ASR & [EN\_US]  ALL \:\: \:\:\:\:YOU \:\: \:\:\:\:NEED\\
LID$_{\mathsf{tok}}$ & [EN\_US] [EN\_US] [EN\_US] [EN\_US] \\
LID$_{\mathsf{utt}}$ & [EN\_US]\\
\hline
\end{tabular}
\end{table}

\noindent We then create two different sets of labels that can represent $I$: utterance-level LIDs (LID$_{\mathsf{utt}}$) and token-level LIDs (LID$_{\mathsf{tok}}$). 
An utterance-level LID is a single label from the set of all possible languages $i^{\mathsf{utt}} \in \mathcal{B}$.
In other words, only a single LID token is used as the ground truth. 
Alternatively, we can define a $S$-length token-level LID sequence corresponding to each $S$-length label sequence as follows: $I^{\mathsf{tok}}=\{i_s^\mathsf{tok} \in \mathcal{B}|s=1,...,S\}$.
LID$_{\mathsf{tok}}$ thus explicitly aligns the language with each spoken word. This approach, inspired by code-switching \cite{byanCS}, forces the model to predict both the frame-level alignment and segmentation between tokens. The task effectively becomes one of identifying the language of each token rather than each utterance. We hypothesize this will aid the model in mapping the audio to the large multilingual text space, even without any code-switched utterances. Example labels can be found in Table (\ref{tab:labels}). 

\vspace{-1.5mm}
\subsection{Hierarchical conditioning}
\label{sec:hier}

\begin{figure}[tb]
    \centering\resizebox{\columnwidth}{!}{
    \includegraphics[]{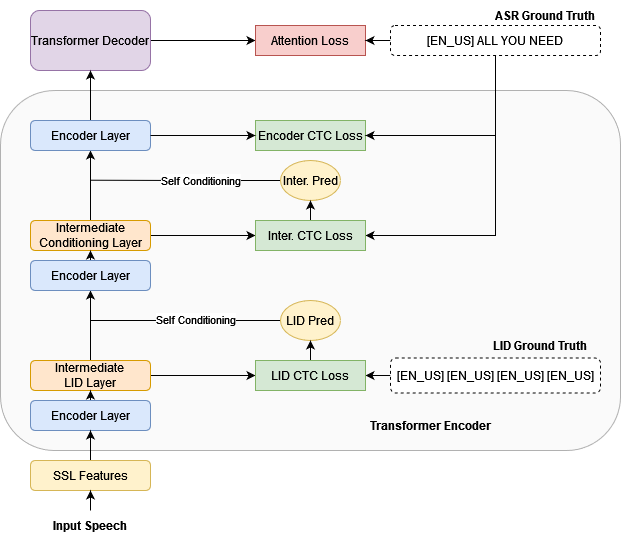}}
    \caption{Proposed hierarchical architecture. The LID predictions of the intermediate layer are used to train the next layer.}
    \label{fig:my_label}
\end{figure}

Explicitly training all intermediate layers on LID allows the model to condition on language information, but perhaps early layers may be sufficient to predict LID, allowing later layers to predict intermediate transcripts instead. This progression can be realized using hierarchical conditioning \cite{higuchihier,sanabria2018hierarchical, yan2022ctc}, where layers perform incrementally more complex predictions. We construct a hierarchical setup of $K$ intermediate layers, such that the $k=1$ intermediate layer is trained using Equation (\ref{eq:lid_loss}) to predict the spoken language (Figure \ref{fig:my_label}). The auxiliary LID task is given to an earlier intermediate layer, such that following encoder layers can be conditioned on its prediction. Later intermediate layers are trained with SC-CTC to keep the regularization benefits. 

\begin{align}
    \mathbf{h^\text{int}} &= \textsc{Enc}^{\text{int}}(\textsc{Nrm}(\mathbf{h^\text{lid}}) + \textsc{Lin}(Z^\text{lid}))) \label{eq:hier_add} \\
        \mathbf{h} &= \textsc{Enc}^{\text{fin}}(\textsc{Nrm}(\mathbf{h^\text{int}}) + \textsc{Lin}(Z^\text{int}))) \label{eq:hier_add2} \\
    P_\text{CTC}(Z | X, Z^{\text{lid}}) &= \textsc{CTC}(\mathbf{h}) \label{eq:hier_ctc}
\end{align}
The encoder output $\mathbf{h}$ is therefore both self-conditioned and LID-conditioned. Equation \ref{eq:lid_loss} can be summed with the loss of all SC-CTC layers, which is then averaged to produce the objective function used to train the intermediate layers:
\begin{equation}
    \mathcal{L}^{\mathsf{hier}} = \frac{1}{K}( \mathcal{L}^{\mathsf{lid}} + \sum_{k=2}^{K}\mathcal{L}^{\mathsf{inter}}_k)
    \label{eq:hc_lid_loss}
\end{equation}
Where $\mathcal{L}^{\mathsf{inter}}$ is the negative log-likelihood of Equation (\ref{eq:P_inter}), the posterior CTC distribution of an intermediate layer. The overall CTC loss can then be obtained in Equation (\ref{loss-ctc-overall}) with a weighted sum of the hierarchical loss (Equation \ref{eq:hc_lid_loss}) with the CTC loss of the full encoder network, where $w$ is the weight of the intermediate losses.

\begin{equation}
    \mathcal{L}^\textsc{CTC} = (1-w) \mathcal{L}^{\textsc{CTC}}_\mathsf{enc}+ w \mathcal{L}^\mathsf{hier}
    \label{loss-ctc-overall}
\end{equation}

\noindent Substituting Equation (\ref{loss-ctc-overall}) into Equation (\ref{loss_hybrid}) yields the complete loss function used to train our encoder-decoder network with both LID conditioning and SC-CTC:

\begin{equation}
    \mathcal{L} = (1-\lambda) \mathcal{L}^\mathsf{att} + \lambda((1-w) \mathcal{L}^{\textsc{CTC}}_\mathsf{enc}+ w \mathcal{L}^\mathsf{hier})
    \label{loss_breakdown}
\end{equation}

\noindent Specifically, our model jointly optimizes the encoder with the LID-conditioned intermediate loss, the CTC loss, and the encoder-decoder attention loss. The decoder is conditioned on the prepended LID token and optimized with the attention loss (Figure~\ref{fig:my_label}).
\vspace{-1.5mm}
\section{Experiments}
 \vspace{-1.5mm}

\subsection{Datasets}
As discussed in Sec.~\ref{sec:intro}, the experiments were conducted on FLEURS \cite{conneau2022fleurs}, a 102-language ASR dataset. Each utterance is a news snippet read by a speaker and contains only one language. Each language in FLEURS has around 7-10 hours of training data, for a total of 987 training hours. Due to the limited amount of supervised data for each language, we experimented with two pre-trained Self-Supervised Learning (SSL) features that performed well on SUPERB \cite{yang21c_superb}: XLS-R \cite{babu2021xls} and WavLM \cite{chenWavlm}. The acoustic inputs are augmented by SpecAugment~\cite{park2019specaugment} and speech perturbation \cite{ko2015audio}. Input text was prepended by language identification tokens, before tokenization by SentencePiece~\cite{kudo2018sentencepiece} with a vocabulary size of 6500. 

\subsection{Model Configuration}
All experiments were conducted through ESPnet2 \cite{watanabe2018espnet}. We use an encoder-decoder setup trained on the hybrid CTC/Attention \cite{watanabe2017Hybrid} multi-task objective, with a CTC weight  of $\lambda=0.3$. We experiment with both Transformer \cite{vaswani2017attention} and Conformer \cite{gulatiConformer} architectures as the encoder. The encoder has either 18 Transformer layers or 12 Conformer layers. Each encoder layer has 8 attention heads and 2048 hidden units. The 6-layer Transformer decoder also has 8 attention heads and 2048 hidden units each. We average 3 checkpoints with the highest validation accuracy. We perform joint CTC/Attention decoding with a language model, using a beam size of 10 and CTC weight of 0.3. Model parameters totaled to around 102 million. 

\vspace{-1.5mm}
\subsubsection{Baseline Models}
\noindent \textbf{CTC/Attention:} A hybrid CTC/Attention model trained multilingually without any intermediate CTC objectives.

\noindent \textbf{SC-CTC:} A model trained with intermediate self-conditioned CTC~\cite{nozaki2021relaxing}, as discussed in Sec.~\ref{sec:sc-ctc}. The intermediate label is identical to the ASR ground truth. We use the same Transformer SC-CTC parameters as~\cite{nozaki2021relaxing}: 5 intermediate layers (3, 6, 9, 12, 15) with an intermediate CTC weight of $w=0.5$. For the 12-layer Conformer encoder, we use intermediate layers 3,6, and 9.

\vspace{-1.5mm}
\subsubsection{Proposed Models}

\noindent \textbf{LID$_{\mathsf{utt}}$ \& LID$_{\mathsf{tok}}$:} Models trained with the proposed intermediate tasks that explicitly leverage the LID described in Sec.~\ref{sec: lid-cond}. The intermediate layer configuration is the same as SC-CTC. In the LID$_{\mathsf{utt}}$ model, all intermediate layers use a single LID token as the output label. For LID$_{\mathsf{tok}}$, the ground truth is comprised of an LID token for each token in the original utterance.

\noindent \textbf{HierLID$_{\mathsf{utt}}$ \& HierLID$_{\mathsf{tok}}$:} Our proposed model that incorporates the LID prediction task into a hierarchical setup (Sec. \ref{sec:hier}). The first intermediate layer (layer 3) uses the LID as the CTC objective, while deeper intermediate layers (6,9,12,15) use the ASR text. We report results for both LID$_{\mathsf{utt}}$ and LID$_{\mathsf{tok}}$ as the first objective.

\vspace{-1.5mm}
\section{Results}
\vspace{-1.5mm}

\begin{table}
\centering
\caption{Comparing the effectiveness of SSL features, reporting CER, MER, LID \% accuracy on FLEURS. XLS-R significantly outperforms WavLM in multilingual ASR.}
\label{tab:result_ssl}
\resizebox{\columnwidth}{!}{
\begin{tabular}{clcccc}
\hline
\texttt{ID} & Model & SSL Features & \multicolumn{3}{c}{Test}\\ 
 & & & CER($\downarrow$) & MER($\downarrow$) & LID($\uparrow$) \\ \hline
\multicolumn{5}{l}{\textit{Transformer}} \\
\texttt{A1} & CTC/Attention & WavLM & 14.6 & 41.8 & 95.09 \\
\texttt{A2} & +SC-CTC & WavLM &  14.4 & 40.8 & 94.47 \\
\texttt{B1} & CTC/Attention & XLS-R & 13.9 & 39.7 & \textbf{95.73}\\
\texttt{B2} & +SC-CTC & XLS-R &  \textbf{13.7} & \textbf{38.8} & 95.39\\ 
 \hline
\end{tabular}}
\vskip -0.2in
\end{table}

We report both Character Error Rate (CER) and Mixed Error Rate (MER), along with the language identification accuracy (LID). MER is calculated using the CER for languages not delimited by white space, and Word Error Rate (WER) for all other languages. Table \ref{tab:result_ssl} shows our early experiments with different pre-trained SSL models. While self-conditioning improved the results of both models (\texttt{A1} vs. \texttt{A2}, \texttt{B1} vs. \texttt{B2}), XLS-R consistently outperformed WavLM and achieved SOTA performance. This result was apparent in early development, so we did not continue experimentation with WavLM.

\begin{table}[tb]
\centering
\caption{
Character error rate (CER), mixed error rate (MER),  and language identification \% accuracy (LID) on FLEURS.
}
\label{tab:result}
\begin{tabular}{clccc}
\hline
\texttt{ID} & Model  & \multicolumn{3}{c}{Test}\\ 
  & & CER($\downarrow$) & MER($\downarrow$) & LID ($\uparrow$)\\ \hline
 \multicolumn{5}{l}{\textit{Prior Work}} \\
\texttt{Z1} & w2v-bert-51 \cite{conneau2022fleurs} & 14.1 & - & - \\
\texttt{Z2} & mSLAM-101 \cite{conneau2022fleurs} & 14.6 & - & - \\
\hline\hline
\multicolumn{5}{l}{\textit{Transformer}} \\
\texttt{B1} & CTC/Attention & 13.9 & 39.7 & \textbf{95.73}\\
\texttt{B2} & +SC-CTC  & \textbf{13.7} & \textbf{38.8} & 95.39\\ 
 \hline
\texttt{C1} & +LID$_{\mathsf{utt}}$   & 13.6 & 37.2 &  95.62 \\ 
\texttt{C2} & +LID$_{\mathsf{tok}}$   &  13.4 & \textbf{35.8} &  \textbf{95.86} \\ %
\texttt{C3} & +HierLID$_{\mathsf{utt}}$   & 13.3 & 36.1 &  95.43 \\ %
\texttt{C4} & +HierLID$_{\mathsf{tok}}$   & 13.3 & 36.0 &  95.31 \\ %
 \hline
  \hline
  \multicolumn{5}{l}{\textit{Conformer}}\\
\texttt{D1} & +SC-CTC   & 10.4 & 32.9  & \textbf{95.41}  \\ 
\texttt{D2} & +HierLID$_{\mathsf{utt}}$  & \textbf{10.1} & \textbf{31.5} & 94.92 \\ 
  \hline
 \hline
\end{tabular}
\vskip -0.2in
\end{table}

\begin{table}[] 
\centering
\caption{Languages with largest differences in  Character Error Rate (CER) ($\downarrow$) between HierLID$_{\mathsf{utt}}$ Conformer and w2v-bert: Georgian (Ka), Cantonese (Yue), Hebrew (He), Swedish (Sv), and Umbundu (Umb).} 
\label{tab:fleurs-cer}
\resizebox{\columnwidth}{!}{
\begin{tabular}{clccc|ccc}
\hline
\texttt{ID} & Model & Ka & Yue & He & Oc & Sv & Umb \\ \hline
\texttt{Z1} & w2v-bert-51 \cite{conneau2022fleurs} & 30.7 & 37.0 & 37.2 & \textbf{11.7} & \textbf{7.6} & \textbf{13.1} \\
\texttt{Z2} & mSLAM-101 \cite{conneau2022fleurs} & 31.0 & 39.8 & 42.5 & 12.7 & 7.8 & 14.0 \\  \hdashline
\texttt{D1} & SC-CTC & \textbf{8.0} & 15.4 & 18.1 & 14.4 & 11.7 & 23.7 \\
\texttt{D2} & HierLID$_{\mathsf{utt}}$ & 8.1 & \textbf{15.3} & \textbf{17.0} & 17.6 & 15.7 & 22.4 \\
 \hline
\end{tabular}}
\vskip -0.2in
\end{table}

Table \ref{tab:result} presents our main results in four partitions: 1) prior work, 2) Transformer baselines, 3) Transformers with the proposed methods, and 4) extended studies with Conformers. Our baseline (\texttt{B1}) improves upon previous works (\texttt{Z1} and \texttt{Z2}) by using XLS-R SSL features \cite{babu2021xls} with a CTC/Attention architecture. Conditioning on both LID and transcriptions further improves ASR performance (\texttt{B1} vs \texttt{B2}). Moreover, explicitly conditioning on the LID is more beneficial than self-conditioning (\texttt{B2} vs. \texttt{C1}, \texttt{C2}). Specifically, LID$_{\mathsf{tok}}$ is more effective than LID$_{\mathsf{utt}}$ (\texttt{C1} v.s. \texttt{C2}); the former even outperforms SC-CTC by 3.0 MER absolute (\texttt{B2} vs. \texttt{C2}). The addition of hierarchical conditioning, however, shrinks this gap (\texttt{C3} vs \texttt{C4}). The combination of both LID$_{\mathsf{utt}}$ and SC-CTC improves over solely LID$_{\mathsf{utt}}$-conditioning by a large degree (\texttt{C1} vs. \texttt{C3}), suggesting that some amount of token-level conditioning is necessary to take advantage of the technique. 

We further push ASR performance by applying these methods to the Conformer. All Conformer models outperform their Transformer variants, and HierLID$_{\mathsf{utt}}$ maintains its advantage over SC-CTC (\texttt{D1} vs. \texttt{D2}). However, due to the increased training instability of the Conformer \cite{guoConformer}, the other methods do not converge with the same optimization settings. Therefore, due to this difference in training stability and the similar performance of the proposed methods in our Transformer trials, we prefer evaluating HierLID$_{\mathsf{utt}}$ (\texttt{D2}) when training Conformer models. The combination of HierLID$_{\mathsf{utt}}$ and the Conformer yields our best result (\texttt{D2}), which outperforms the CER of previous work in equivalent settings by a wide margin: 4.0 absolute\footnote{One concurrent work \cite{chen2022maestro} further improves CER by 1.4, albeit with additional training data \cite{ardila-etal-2020-common, gales14_babel, wang-etal-2021-voxpopuli, pratap20_interspeech}, while another \cite{radfordrobust} was evaluated zero-shot on a subset of languages.}. 

\subsection{Analysis}

To better understand effectiveness of our technique, we conducted an analysis of our results by language. Table \ref{tab:fleurs-cer} compares the best/worst performing languages by HierLID$_{\mathsf{utt}}$ Conformer (\texttt{D2}) relative to w2v-bert (\texttt{Z1}), which can vary as much as 22.7 CER.  These large discrepancies are likely derived from differences in SSL pre-training. Compared to w2v-bert (600M parameters), XLS-R (300M parameters) was pretrained on an additional 6.6K hours of data (436K total) that extended its language coverage by 77. We suspect that the larger parameter size and smaller pool of languages allowed w2v-bert to learn better representations in the languages that it covered, which carried over to ASR. Similarly, Table \ref{tab:fleurs-lid} compares the languages with the largest change in LID accuracy between our two Conformer models. We found that degradations in LID accuracy were often caused by confusion with a related language. However, this was generally accompanied by improvements in the other language, such as with the case of Serbian and Bosnian. In extreme cases, misclassifications considerably affected CER, such as for Swedish and Occitan (Tables \ref{tab:fleurs-cer} and \ref{tab:fleurs-lid}), which were frequently misidentified as Norwegian and French respectively.

\begin{table}[t] 
\vskip -0.2in
\centering
\caption{Languages with  largest differences in LID accuracy ($\uparrow$) between HierLID$_{\mathsf{utt}}$ and SC-CTC Conformer: Zulu (Zu), Hindi (Hi), Bosnian (Bs), Occitan (Oc), Swedish (Sw), and Umbundu (Umb).} 
\label{tab:fleurs-lid}
\resizebox{\columnwidth}{!}{
\begin{tabular}{clccc|ccc}
\hline
\texttt{ID} & Model & Zu & Hi & Bs & Oc & Sv & Umb \\ \hline
\texttt{D1} & SC-CTC & 66.8 & 80.4 & 32.1 & \textbf{48.1} & \textbf{95.3} & \textbf{91.7} \\
\texttt{D2} & HierLID$_{\mathsf{utt}}$ & \textbf{83.6} & \textbf{91.4} & \textbf{42.9} & 35.4 & 75.9 & 60.0 \\
 \hline
\end{tabular}}
\end{table}

\begin{table}[t] 
\vskip -0.2in
\centering
\caption{Average Conformer CER ($\downarrow$) compared to prior work for each language group.}
\label{tab:result_group}
\resizebox{\columnwidth}{!}{
\begin{tabular}{clccccccc}
\hline
\texttt{ID} & Model & WE & EE & CMN & SSA & SA & SEA & CJK \\ \hline
\texttt{Z1} & w2v-bert-51 \cite{conneau2022fleurs} & 10.7 & 9.9 & 14.5 & 15.6 & 17.4 & 14.7 & 25.0 \\
\texttt{Z2} & mSLAM-101 \cite{conneau2022fleurs} & 10.6 & 10.0 & 14.8 & 16.4 & 19.2 & 14.9 & 24.6 \\ \hdashline
\texttt{D1} & SC-CTC & \textbf{9.0} & 7.5 & \textbf{9.1} & 12.6 & 16.3 & 14.6 & \textbf{17.9}\\
\texttt{D2} & HierLID$_{\mathsf{utt}}$ & 9.3 & 7.5 & 9.2 & \textbf{12.0} & \textbf{15.5} & \textbf{13.5} & 18.3 \\
 \hline
\end{tabular}}
\caption{Average Conformer LID \% accuracy ($\uparrow$) compared to prior work for each language group.}
\label{tab:lid_group}
\resizebox{\columnwidth}{!}{
\begin{tabular}{clccccccc}
\hline
\texttt{ID} & Model & WE & EE & CMN & SSA & SA & SEA & CJK \\ \hline
\texttt{Z1} & w2v-bert-51 \cite{conneau2022fleurs} & 85.3 & 78.4 & 72.9 & 59.1 & 52.0 & 65.7 & 89.7 \\
\texttt{Z2} & mSLAM-101 \cite{conneau2022fleurs} & 84.6 & 81.3 & 75.9 & 62.2 & 51.7 & 73.4 & 87.8 \\  \hdashline
\texttt{D1} & SC-CTC & \textbf{94.1} & \textbf{95.1} & \textbf{98.9} & \textbf{96.6} & 89.6 & 94.1 & \textbf{99.3}\\
\texttt{D2} & HierLID$_{\mathsf{utt}}$ & 92.5 & 94.2 & 97.7 & 96.4 & \textbf{90.5} & \textbf{95.4} & 98.9 \\
 \hline
\end{tabular}}
\vskip -0.2in
\end{table}

We also performed a region-level analysis. Table \ref{tab:result_group} shows the CERs for each group in FLEURS: Western Europe (WE), Eastern Europe (EE), Central-Asian, Middle-East and North-Africa (CMN), Sub-Saharan Africa (SSA), South Asia (SA), South-East Asia (SEA), and East Asia (CJK). Both Conformer models improve across-the-board compared to prior work \cite{conneau2022fleurs}, with notable CER reductions in the CJK and CMN language groups. The HierLID$_{\mathsf{utt}}$ technique is particularly effective on the SSA, SA, and SEA language groups compared to SC-CTC, with a small performance cost in WE, CMN, and CJK (\texttt{D1} vs. \texttt{D2}). Table \ref{tab:lid_group} makes a similar comparison using LID accuracy across language groups. Both Conformer models again out-perform previous work across all language groups, but the LID accuracy of HierLID$_{\mathsf{utt}}$ degrades in all but two language groups when compared to SC-CTC (\texttt{D1} vs. \texttt{D2}).

\vspace{-3mm}
\section{Conclusion}
\vspace{-1.5mm}
Improving multilingual ASR can help extend speech technologies to new languages. However, these models face the challenge of handling the typological diversity of so many languages. To help handle this, we introduce a framework using hierarchical CTC that can leverage language identity throughout the entire encoder-decoder network, hypothesizing that correctly identifying the language eases transcription modelling. We evaluate our technique on the 102-language FLEURS dataset to show its effectiveness and improve over the results of prior work. In the future, we hope to extend our approach to an even larger set of languages \cite{li2k_interspeech} and data \cite{Hou2020}, so that these trained models can also in downstream tools, such as with speech alignment and data cleaning, that can further help extend speech technologies to more languages.

\section{Acknowledgements}
This work used the Bridges2 system \cite{nystrom2015bridges}, supported by NSF award number ACI-1445606, at the Pittsburgh Supercomputing Center.
\clearpage

\section{References}
\printbibliography
\end{document}